# THE APPLICATION OF BAYES YING-YANG HARMONY BASED GMMS IN ON-LINE SIGNATURE VERIFICATION

Xiaosha Zhao   Mandan Liu

Key Laboratory of Advanced Control and Optimization for Chemical Processes, Ministry of Education, East China Uni. of Sci. and Tech.

*ABSTRACT*

*In this contribution, a Bayes Ying-Yang(BYY) harmony based approach for on-line signature verification is presented. In the proposed method, a simple but effective Gaussian Mixture Models(GMMs) is used to represent for each user's signature model based on the prior information collected. Different from the early works, in this paper, we use the Bayes Ying Yang machine combined with the harmony function to achieve Automatic Model Selection(AMS) during the parameter learning for the GMMs, so that a better approximation of the user model is assured. Experiments on a database from the First International Signature Verification Competition(SVC 2004) confirm that this combined algorithm yields quite a satisfactory result.*

## 1. INTRODUCTION

The on-line signature verification task can be expressed as follows: given a complete on-line signature $U$ and a claimed user $c$, decide whether $c$ indeed produced the signature $U$ [1]. To implement this procedure mathematically, a model $\Theta^c$ for the user c and $\Theta^-$ an antithetical model need to be learnt so that a score function $S(U,\Theta^c,\Theta^-)$ can be calculated, which will later be used to compare against some pre-set threshold T to finally determine the authenticity of the signature $U$:

$$S(U,\Theta^c,\Theta^-) \begin{cases} \geq T\, accept\, identity\, claim \\ < T\, reject\, identity\, claim \end{cases} \quad (1)$$

Based on the above theory, selecting a good model is the most important step in designing a signature verification system. Despite the most commonly used, distance based Dynamic Warping(DW)[2], or the feature-based statistical method, Hidden Markov Modeling(HMM)[3][4], in this paper, we propose a new BYY based GMMs to build the signature models for the users. The GMMs based recognizers are conceptually less complex than HMM, which leads to significantly shorter training time as well as less parameters to learn[5]. Distinguished from the earlier works with the cluster numbers pre-settled and same for all the users[6], the BYY based GMMs can decide the optimal cluster numbers automatically according to the data distribution of different users in the process of parameter leaning, such that improves the performance of the algorithm.

This paper is structured as follows: the introduction involves some definitions, and the brief description of the main idea in this work. In Section 2, the subset of the BYY based GMMs method we focused on is detailed. Section 3 presents the feature select and data processing used in our work, followed by the model training and similarity score computation in Section4. And the experiment result as well as the performance evaluation of this method proposed is explained in section 5.

## 2. BYY BASED GMMS FOR SIGNATURE VERIFICATION

### 2.1 Gaussian Mixed Models

GMMs are such well known and so much referenced statistical models in many pattern recognition applications. Based on the representation of a weighted linear combination of Gaussian probabilistic function, as shown in the following equation (2),they are versatile modeling tools to approximate any probability density function(pdf) given a sufficient number of components while impose only minimal assumptions about the modeled random variables.

$$p(u|\Theta) = \sum_{j=1}^{k} \alpha_j p(u|\Theta_j)$$

$$p(u|\Theta_j) = G(u|m_j, \sigma_j) = \frac{1}{(2\pi)^{d/2}|\sigma_j|^{1/2}} e^{-\frac{1}{2}(u-m_j)^T \sigma_j^{-1}(u-m_j)} \quad (2)$$

Where $u \in R^d$ are the feature vectors that represent a handwritten signature, $\Theta = \alpha \cup \{\theta_j\}_{j=1}^{k}$, $\alpha = [\alpha_1, \ldots, \alpha_k]^T$, $\theta_j = (m_j, \sigma_j)$, $\alpha_j$ is the mixing weight for the $j$th component with each $\alpha_j \geq 0$ and $\sum_{j=1}^{k} \alpha_j = 1$, and $G(u|m_j, \sigma_j)$ denotes a Gaussian density with a mean $m_j$ and a covariance matrix $\sigma_j$. Each $p(u|\Theta_j)$ is called a component, and k refers to the component number, i.e. the cluster number.

To learn the unknown parameter Θ, the EM algorithm has been used in an iterative mode. However, one defect bothering in this method is that the component number k needs to be set manually, which when not consist with the actual data distribution, will leads to local optimum and further impact the accuracy of the verification. Some prior works would pick several promising values to run and choose the best one as k. In addition to the additional time cost, the discontinuity of the values in such method can also miss the best choice for k. The BYY, on the other hand, can choose the optimal cluster number automatically during the parameter learning.

## 2.2 The Bayes Ying-Yang theory and harmony function

A BYY system describes each data vector $u \in U \in R^d$ and its corresponding inner representation $y \in \mathcal{Y} \in R^d$ using two types of Bayesian decomposition of the joint density $p(u,y) = p(y|u)p(u)$ and $p(u,y) = q(u|y)q(y)$, named as Yang machine and Ying machine, respectively[7]. Given a data set $U_r = \{u_t\}_{t=1}^N$, where N is the total number of the sample, from the observation space $U$, the task of learning on a BYY system is mainly to specify each of $p(y|U_r)$, $p(U_r)$, $q(U_r|y)$, $q(y)$ with a harmony learning principle implemented by maximizing the functional[8]:

$$H(p||q) = \int p(y|U_r)p(U_r) \ln[q(U_r|y)q(y)] \, dU_r dy - \ln z_q \qquad (3)$$

where $z_q$ is a regulation term.

As a matter of fact, the maximization of the harmony function $H(p||q)$ can push to get best parameter match as well as the least structure complexity, thus to produce the favorite property of AMS as long as k is set to be larger than the true number of components in the sample data $U_r$. Based on this algorithm, it can be applied on the GMMs learning to strive for better accuracy.

## 3. FEATURE SELECT AND DATA PROCESSING

The signature sample data employed in our work are sampled by the pen tablets, which can detect the horizontal position($x_t$), vertical position($y_t$), pressure($p_t$) and azimuth($a_t$) of the pen point, as well as the elevation of the pen. Besides, the sensor also records the pen-up($p_t = 0$) ($z_t$)points. So the raw signature vector can be expressed as follows:

$$\widetilde{u_t} = [x_t, \, y_t, \, p_t, \, a_t, \, z_t] \qquad (4)$$

Referring to former experimental examples[2], as well as considering the practical application, we restricted the investigation to horizontal position, vertical position and pressure data. Besides, to get more discriminative, two dynamic features, trajectory tangent angle $\theta_t$ and instantaneous velocity $v_t$, which two are difficult to reproduce based only on visual inspection[9], are computed as follows:

$$\theta_t = \arctan\frac{\dot{y}_t}{\dot{x}_t} \quad v_t = \sqrt{\dot{x}_t^2 + \dot{y}_t^2} \qquad (5)$$

where $\dot{x}_t$, $\dot{y}_t$ represent the first derivatives of $x_t$ and $y_t$ with respect to time. Finally, we get the basic feature vector for each sample:

$$u_t' = [x_t, \, y_t, \, p_t, v_t, \theta_t] \qquad (6)$$

Figure 1 gives an example of the signature date from one user.

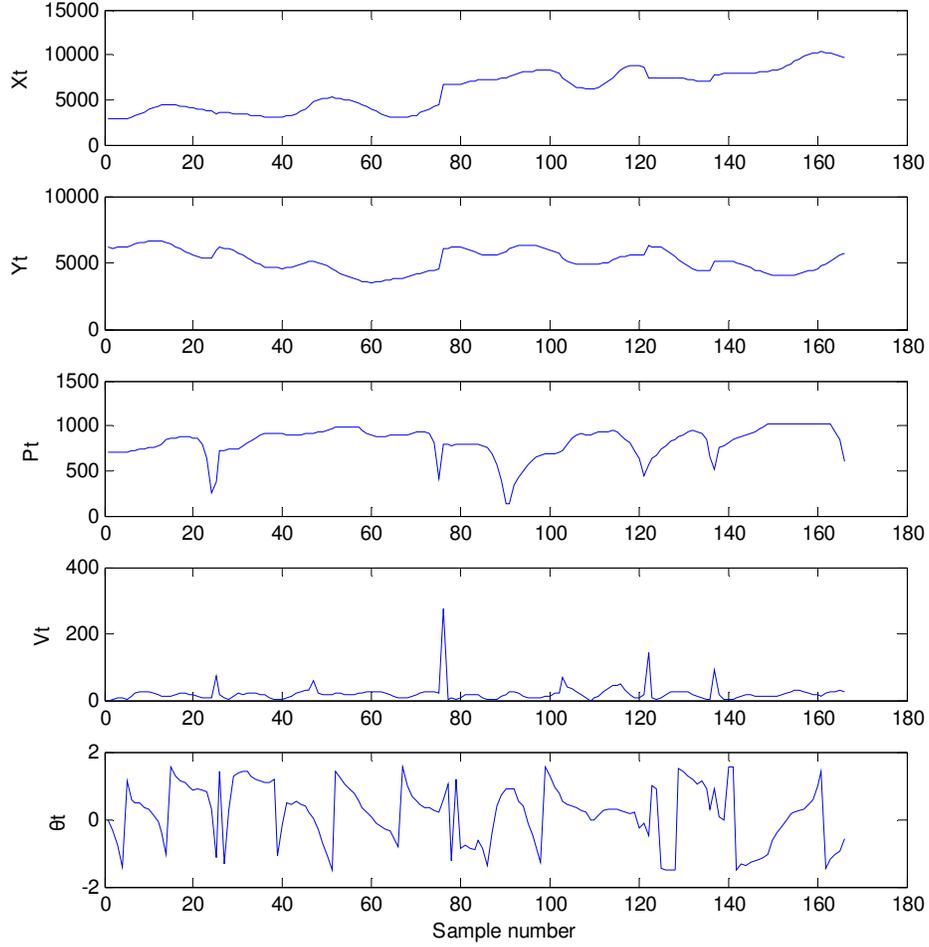

Figure 1 Signature data

To eliminate the dynamic ranges of the different features and ensure a better learning result, each individual feature $u_{dt}' \in \{x_t,\ y_t,\ p_t,\ \theta_t,\ v_t\}$ with $d$=1,...,5 is transformed into a zero-mean, unit variance normal distribution using:

$$u_{dt} = \frac{u_{dt}' - u_d'}{\sigma_d'} \quad (7)$$

Where $u_d'$ is the mean value of the $dth$ dimension vectors, and $\sigma_d'$ the corresponding variance value. After the transformation, we can get the unified signature vector $u_t$:

$$u_t = [\bar{x}_t,\ \bar{y}_t,\ \bar{p}_t, \bar{v}_t,\ \bar{\theta}_t] \quad (8)$$

And the final complete observation comes as

$$U = [U_1 \ldots, U_r \ldots, U_M] \tag{9}$$

where M is the total number of the users.

The complete verification process is described in the following flow chart:

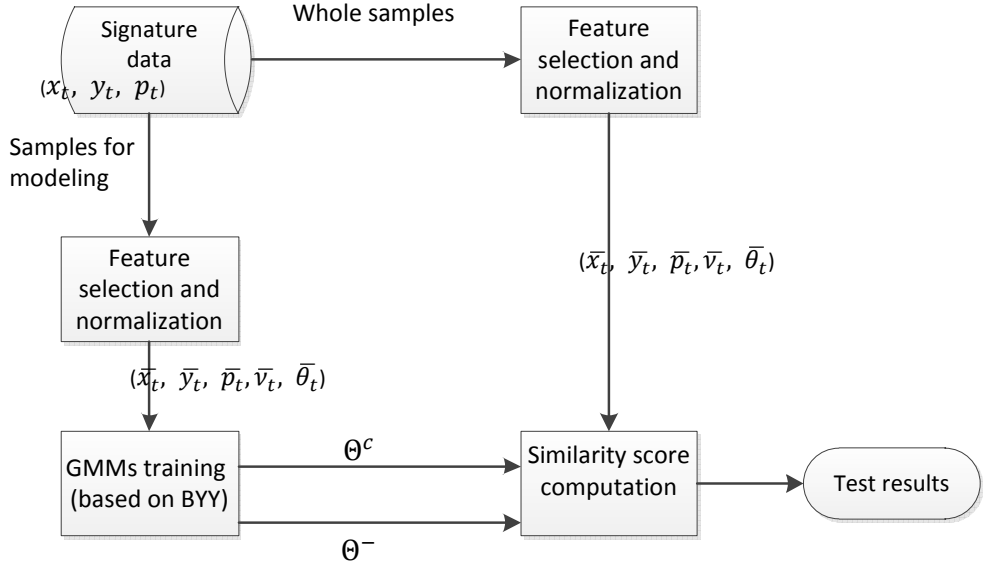

Figure 2 The complete verification follow chart

## 4 MODEL TRAINING AND SIGNATURE SCORE COMPUTING

### 4.1 Model training

The main task of the model training is to maximize the harmony function $H(p||q)$. And in our work, we chose annealing learning algorithm proposed in [10], which sets the regulation term $z_q$ in equation (3) to one, and $p(U_r)$ to some empirical density estimation:

$$p(U_r) = \frac{1}{N}\sum_{t=1}^{N} K(u - u_t) \tag{9}$$

where K(.) is a prefixed kernel function[8], and further converge to the delta function:

$$K(u - u_t) = \begin{cases} +\infty, & u = u_t \\ 0, & u \neq u_t \end{cases} \tag{10}$$

.According to Bayes' law and the definition of GMMs:

$$p(y = j|U_r) = \frac{\alpha_j q(U_r|\theta_j)}{q(U_r|\Theta_k)}, \quad q(U_r|\Theta_k) = \sum_{j=1}^{k} \alpha_j q(U_r|\theta_j) \tag{11}$$

where $q(U_r|\theta_j) = q(U_r|y=j), \theta_j$ represents the unknown parameters in each component and $\Theta_k = \{a_j, \theta_j\}_{j=1}^k$ represents all the parameters.

Substituting (9)-(11) into (3), can we get :

$$L(\Theta_k) = H(p||q) = \frac{1}{N}\sum_{t=1}^{N}\sum_{j=1}^{k}\frac{a_j q(u_t|\theta_j)}{\sum_{i=1}^{k} a_i q(u_t|\theta_i)} \ln[a_j q(u_t|\theta_j)] \qquad (12)$$

As we consume the components as Gaussian functions, so:

$$q(u_t|\theta_j) = q(u_t|m_j, \sigma_j) = \frac{1}{(2\pi)^{d/2}|\sigma_j|^{1/2}} e^{-\frac{1}{2}(u_t-m_j)^T \sigma_j^{-1}(u_t-m_j)} \qquad (13)$$

with the $\Theta_k$ changed into $\Theta_k = \{a_j, m_j, \sigma_j\}_{j=1}^k$, while the $p(y|u_t)$ is a free probability distribution under the basic probability constrains. In this situation, the harmony function can be rewritten as

$$L(\Theta_k) = \frac{1}{N}\sum_{t=1}^{N}\sum_{j=1}^{k} p(j|u_t) \ln[a_j q(u_t|m_j, \sigma_j)] \qquad (14)$$

with the parameters $\Theta_k = \{a_j, m_j, \sigma_j, t=1,\ldots,N\}_{j=1}^k$.

However, one problem here to learn directly on the equation (14) is that the learning result makes it the hard-cut EM algorithm[11], which can be easily trapped in a local maximum while the component number k set bigger than the true one during the training as stated earlier. To get an optimum k, the annealing algorithm attaches a soften item to $L(\Theta_k)$ in (15):

$$L_\lambda(\Theta_k) = \frac{1}{N}\sum_{t=1}^{N}\sum_{j=1}^{k} p(j|u_t)\ln[a_j q(u_t|m_j, \sigma_j)] + \lambda O_N(p(y|U_r)) \qquad (15)$$

where

$$O_N(p(y|U_r)) = -\frac{1}{N}\sum_{t=1}^{N}\sum_{j=1}^{k} p(j|u_t)\ln p(j|u_t) \qquad (16)$$

By controlling $\lambda \to 0$ from $\lambda_0 = 1$, the maximum of $L_\lambda(\Theta_k)$ can lead to the global maximum of the harmony function $L(\Theta_k)$.

The annealing learning algorithm can be realized by alternatively maximizing $L(\Theta_k)$ with $\Theta_1 = \{p(j|u_t), t=1,\ldots,N\}_{j=1}^k$ and $\Theta_2 = \Theta_k$, as shown follows:

$$p(j|u_t) = \frac{[a_j q(u_t|m_j, \sigma_j)]^{\frac{1}{\lambda}}}{\sum_{i=1}^{k}[a_i q(u_t|m_i, \sigma_i)]^{\frac{1}{\lambda}}} \qquad (17)$$

$$a_j^* = \frac{1}{N}\sum_{t=1}^{N} p(j|u_t) \qquad (18)$$

$$m_j^* = \frac{1}{\sum_{t=1}^{N} p(j|u_t)}\sum_{t=1}^{N} p(j|u_t) u_t \qquad (19)$$

$$\Sigma_j^* = \frac{1}{\sum_{t=1}^{N} p(j|u_t)} \sum_{t=1}^{N} p(j|u_t)(u_t - m_j^*)(u_t - m_j^*)^T \qquad (20)$$

At last, we will get the optimum models for each user, which will be used in the following steps.

### 4.2 Signature score computation

As we can get access to both genuine and forgery signature data, the test model for the signature score computation we use here is the ratio of the posterior probabilities. Suppose the prior probability of a forgery is $p_f$, then $1 - p_f$ is a genuine one's prior probability. Let $G(U, \Theta)$ denote the Gaussian density of the model $\Theta$ evaluated at u. Thus the signature score can be expressed in equation (20).

$$S(U, \Theta^c, \Theta^-) = \frac{G(U, \Theta^c)(1-p_f)}{G(U, \Theta^-)p_f} \qquad (20)$$

And according to the Bayes-optimal classification rule, when $S(U, \Theta^c, \Theta^-) < 1$, which means the probability of the test signature U belonging to $\Theta^-$ is bigger than that of $\Theta^c$, so we decide it to be forgery, otherwise genuine[12].However, as the sample users are limited in our work, so we adjust the threshold to 2 to get a better recognition rate.

## 5 EXPERIMENT RESULTS

### 5.1 The performance of the BYY based GMMs

The data used in our experiment is derived from the First International Signature Verification Competition(SVC 2004)[13]. In this experiment, we use 1600 signatures from 40 users, consisting of 20 genuine ones and 20 forgeries of each, which means the $p_f$ in equation (20) to be 0.5. During the experiment, the first 5 out of 20 genuine signatures from each user were used to build the model $\Theta^c$, and the first 5 forgeries were used for the model $\Theta^-$.

As stated above, the BYY based GMMs is able to choose the optimal cluster number k automatically, so different users can have different component number in his/her GMMs $\Theta^c$. Even more, according to our experiment results, the component number for $\Theta^c$ and $\Theta^-$ of the same user can also be different, as shown in the following table 1.

Table 1 The component numbers (k) in $\Theta^c$ and $\Theta^-$ of each user

|  | k in $\Theta^c$ | k in $\Theta^-$ |  | k in $\Theta^c$ | k in $\Theta^-$ |
|---|---|---|---|---|---|
| User1 | 15 | 23 | User21 | 8 | 8 |
| User2 | 5 | 5 | User22 | 8 | 8 |
| User3 | 18 | 20 | User23 | 26 | 34 |
| User4 | 21 | 18 | User24 | 8 | 8 |
| User5 | 8 | 8 | User25 | 18 | 16 |

| User6  | 16 | 16 | User26 | 8  | 8  |
|--------|----|----|--------|----|----|
| User7  | 30 | 17 | User27 | 8  | 8  |
| User8  | 8  | 8  | User28 | 16 | 12 |
| User9  | 24 | 24 | User29 | 30 | 28 |
| User10 | 19 | 12 | User30 | 22 | 22 |
| User11 | 20 | 18 | User31 | 8  | 8  |
| User12 | 24 | 23 | User32 | 16 | 16 |
| User13 | 32 | 32 | User33 | 14 | 6  |
| User14 | 16 | 16 | User34 | 22 | 8  |
| User15 | 24 | 24 | User35 | 8  | 8  |
| User16 | 32 | 32 | User36 | 32 | 32 |
| User17 | 16 | 16 | User37 | 10 | 18 |
| User18 | 32 | 32 | User38 | 16 | 16 |
| User19 | 5  | 5  | User39 | 32 | 32 |
| User20 | 14 | 12 | User40 | 13 | 22 |

Based on the models built in table 1, along with the threshold chosen, we can finally get the signatures recognized. In order to get a whole vision of the recognition results, Figure 1 shows two examples of the logarithm of the similarity scores computed against the threshold in our experiment for User1 and User 5, respectively.

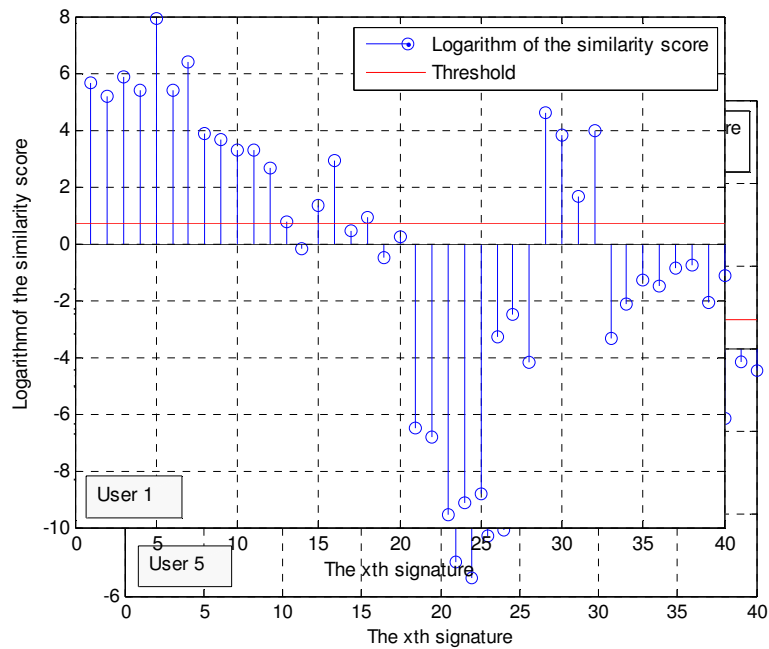

Figure3 The verification results of User 1 and User 5

Table 2 The FAR, FRR and verification rate of each user by BYY based GMMs

|  | FAR(%) | FRR(%) | Verification Rate(%) |  | FAR(%) | FRR(%) | Verification Rate(%) |
|---|---|---|---|---|---|---|---|
| User1 | 10.0000 | 10.0000 | 80.0000 | User21 | 0.0000 | 0.0000 | 100.0000 |
| User2 | 2.5000 | 5.0000 | 92.5000 | User22 | 0.0000 | 0.0000 | 100.0000 |
| User3 | 2.5000 | 0.0000 | 97.5000 | User23 | 0.0000 | 0.0000 | 100.0000 |
| User4 | 5.0000 | 0.0000 | 95.0000 | User24 | 0.0000 | 2.5000 | 97.5000 |
| User5 | 0.0000 | 0.0000 | 100.0000 | User25 | 2.5000 | 0.0000 | 97.5000 |
| User6 | 0.0000 | 0.0000 | 100.0000 | User26 | 0.0000 | 5.0000 | 95.0000 |
| User7 | 0.0000 | 10.0000 | 90.0000 | User27 | 15.0000 | 0.0000 | 85.0000 |
| User8 | 5.0000 | 12.5000 | 82.5000 | User28 | 0.0000 | 2.5000 | 97.5000 |
| User9 | 0.0000 | 0.0000 | 100.0000 | User29 | 0.0000 | 0.0000 | 100.0000 |
| User10 | 2.5000 | 0.0000 | 97.5000 | User30 | 0.0000 | 0.0000 | 100.0000 |
| User11 | 0.0000 | 5.0000 | 95.0000 | User31 | 0.000 | 0.0000 | 100.0000 |
| User12 | 0.0000 | 2.5000 | 97.5000 | User32 | 20.0000 | 0.0000 | 80.0000 |
| User13 | 0.0000 | 0.0000 | 100.0000 | User33 | 12.5000 | 0.0000 | 87.5000 |
| User14 | 0.0000 | 0.0000 | 100.0000 | User34 | 0.0000 | 2.5000 | 97.5000 |
| User15 | 7.5000 | 5.0000 | 87.5000 | User35 | 7.5000 | 0.0000 | 92.5000 |
| User16 | 0.0000 | 12.5000 | 87.5000 | User36 | 0.000 | 12.5000 | 87.5000 |
| User17 | 7.5000 | 0.0000 | 92.5000 | User37 | 15.0000 | 5.0000 | 80.0000 |
| User18 | 0.0000 | 0.0000 | 100.0000 | User38 | 5.0000 | 7.5000 | 87.5000 |
| User19 | 0.0000 | 0.0000 | 100.0000 | User39 | 0.0000 | 0.0000 | 100.0000 |
| User20 | 0.0000 | 0.0000 | 100.0000 | User40 | 0.0000 | 0.0000 | 100.0000 |

And the False Reject Rate(FRR), the False Accept Rate(FAR) as well as the recognition rate are listed in the following table 2. From table 2 we can see that the BYY based GMMs can achievean average recognition rate of 94.5000%, with FAR at 3.0000% and FRR at 2.5000%.

## 5.2 Comparison with the traditional GMMs and DTW

To evaluate the performance of the BYY based GMMs, we use two other recognition methods, the traditional GMMs and DTW, based on the same signature data base.

In the traditional GMMs experiment, the feature vector, the computation of the similarity score as well as the threshold are the same as used in the BYY based GMMs experiment, except that the models are learnt by the EM algorithm. As the component number has to be settled before hand, so we set the k to 8, 16, 24 and 32, respectively, to get the best result. The average FAR, FFR and verification rate are listed in Table 3.

Table 3 The average FAR, FRR and verification rate of the 40 users by normal GMMs

|  | k=8 | k=16 | k=24 | k=32 |
| --- | --- | --- | --- | --- |
| FAR(%) | 6.4375 | 7.7500 | 12.9375 | 14.9375 |
| FRR(%) | 6.5625 | 4.0625 | 5.2500 | 9.1875 |
| Verification Rate(%) | 87.0000 | 88.1875 | 81.8125 | 75.8750 |

From Table 3 it can be concluded that the normal GMMs achieves the best performance with an average verification rate of 88.1875%, FAR at 7.7500% and FRR at 4.0625% when k is set to 16. As to the method of DTW, the feature vector is extracted by way of interpolation and wavelet function, including total sample time, the ratio of height and width, standard deviation in horizontal and vertical direction, standard deviation of pressure, rotation and azimuth, average velocity in horizontal and vertical direction, average pressure, azimuth and rotation, pressure, rotation and azimuth energy extracted by wavelet function, adding up to 36 features altogether.

Among the first 5 genuine signatures, the smallest values in each of the 36 features form a new 36 feature vector $V_s$; and the biggest values in each of the 36 features form another 36 feature vector $V_b$. $V_s$ is used as the model and the matching distance between $V_s$ and $V_b$ calculated by DTW is used as the threshold. The average recognition rate is 68.9375%, the FAR is 17.6875% and the FRR is 13.3750%.

### 5.3 Conclusion

Comparing with the experiment results of traditional GMMs and DTW can we find that the BYY based GMMs has a significant better performance, which proves that the BYY based GMMs can build relatively accurate models for the users, and its application in signature verification produces satisfactory results based on the data samples. So it can be a promising solution in this field.

**Authors**

Xiaosha Zhao (1989-), female, obtained her B.Eng. degree in Measurement and Control Techniques and Instrument from East China University of Science and Technology, in 2012. She is currently a M.D student of Department of Automation in East China University of Science and Technology. Her research interests include statistical learning and clustering.

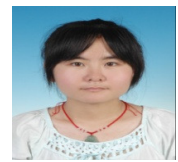

Mandan Liu (1973-), female, Professor, Doctor, vice-president of School of Information Science and Engineering, East China University of Science and Technology, Executive Member of China Academy of System Simulation, majoring in Intelligent Control Theory and Application and Intelligent Optimization Algorithm and Application.

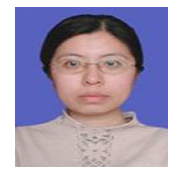